\documentclass[letterpaper]{article} 
\usepackage{float}
\usepackage{booktabs}
\usepackage{aaai24}  
\usepackage{times}  
\usepackage{helvet}  
\usepackage{courier}  
\usepackage[hyphens]{url}  
\usepackage{graphicx} 
\usepackage{natbib}  
\usepackage{caption} 
\usepackage{algorithm}
\usepackage{algorithmic}

\usepackage{newfloat}
\usepackage{listings}
\DeclareCaptionStyle{ruled}{labelfont=normalfont,labelsep=colon,strut=off} 
\floatstyle{ruled}
\newfloat{listing}{tb}{lst}{}
\floatname{listing}{Listing}
\title{\title{Multi-Garment Customized Model Generation}}

\author {
    Yichen Liu,\textsuperscript{\rm 1}
    Penghui Du, \textsuperscript{\rm 2}
    Yi Liu \textsuperscript{\rm 2}
    Quanwei Zhang \textsuperscript{\rm 3}
}
\affiliations {
    \textsuperscript{\rm 1}University of Chinese Academy of Sciences \\
    \textsuperscript{\rm 2}Beihang University \\
    \textsuperscript{\rm 3}ZheJiang University
}

\usepackage{bibentry}

\begin{document}

\maketitle

\begin{abstract}
This paper introduces Multi-Garment Customized Model Generation, a unified framework based on Latent Diffusion Models (LDMs) aimed at addressing the unexplored task of synthesizing images with free combinations of multiple pieces of clothing. The method focuses on generating customized models wearing various targeted outfits according to different text prompts. The primary challenge lies in maintaining the natural appearance of the dressed model while preserving the complex textures of each piece of clothing, ensuring that the information from different garments does not interfere with each other.
To tackle these challenges, we first developed a garment encoder, which is a trainable UNet copy with shared weights, capable of extracting detailed features of garments in parallel. Secondly, our framework supports the conditional generation of multiple garments through decoupled multi-garment feature fusion, allowing multiple clothing features to be injected into the backbone network, significantly alleviating conflicts between garment information. Additionally, the proposed garment encoder is a plug-and-play module that can be combined with other extension modules such as IP-Adapter and ControlNet, enhancing the diversity and controllability of the generated models.
Extensive experiments demonstrate the superiority of our approach over existing alternatives, opening up new avenues for the task of generating images with multiple-piece clothing combinations.
\end{abstract}

\begin{figure*}[htb]
\centering
\includegraphics[width=\linewidth]{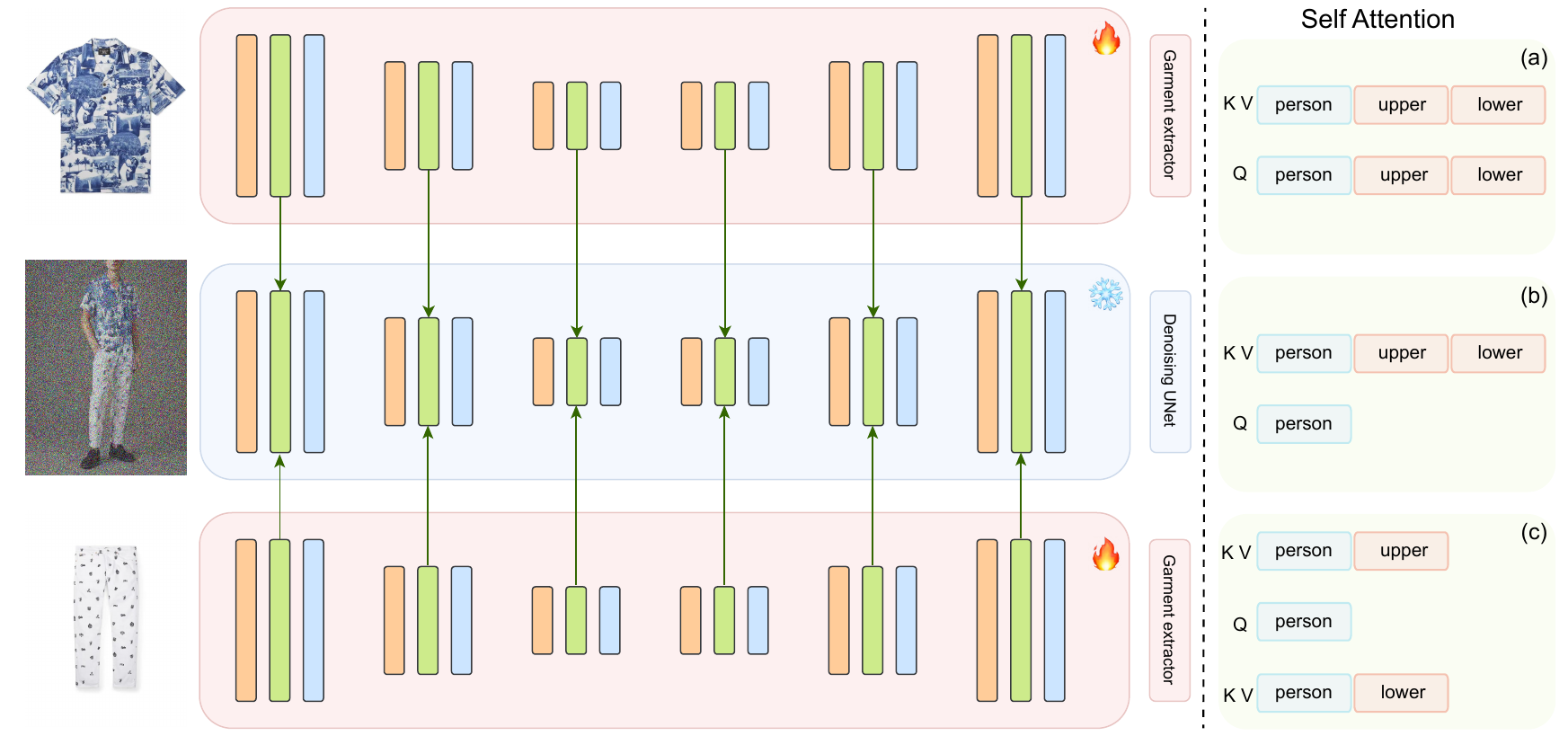}
\caption{The framework of our method.}
\label{fig:1}
\end{figure*}

\section*{1 Introduction}

In recent years, the field of image generation has seen transformative advancements, with Latent Diffusion Model (LDM)-based methods achieving remarkable success in text-to-image generation tasks. Many research efforts have attempted to incorporate control conditions beyond text, such as pose, densepose, sketches, and facial features, for image generation. However, to date, few studies have focused on image generation conditioned on specific clothing. This is a highly promising area, particularly with significant commercial value in e-commerce. Such tasks present two key challenges: on the one hand, adjusting the generated person's pose, background, and facial features through text prompts or additional control conditions; on the other hand, maintaining the texture and style of the target clothing.

Traditional subject-driven image generation methods fall short for these tasks, as they emphasize holistic information of the conditions, such as appearance, structure, and style, but fail to preserve clothing details. So far, research focused on generating models based on given clothing and text prompts (like OMS~\cite{chen2024magic} and StableGarment) has been limited to single garments, unable to meet the commercial need for models showcasing multiple clothing items.

To address this issue, we propose Multi-Garment Customized Model Generation , a network architecture based on LDM that customizes models based on multiple reference garments and text prompts. Inspired by reference mechanisms in image-to-video tasks, we developed a semantic clothing encoder based on the Reference UNet architecture, a trainable UNet replica with shared weights that can extract detailed features of clothing in parallel. Unlike previous attention concatenation operations, we adopt an attention decoupling approach to inject clothing features into the image generation process, preserving the appearance details and spatial information of each garment. Additional semantic conditions represent the categories of the clothing, ensuring different types of garments are not confused during the generation process. Furthermore, our attention reordering operation ensures smooth transitions at the junctions of different clothing items, avoiding abrupt seams. Our semantic clothing encoder is a plug-in module that can be combined with various base models or other extension modules (like ControlNet and IP-Adapter~\cite{ye2023ip-adapter}), specifying the model's pose, facial features, and style while maintaining the details of multiple garments.

With these methods, our model demonstrates high quality and controllability in multi-garment model image generation. Our contributions are summarized as follows:

\begin{itemize}
\item We constructed a novel framework, Multi-Garment Customized Model Generation, supporting text-prompted, multi-garment controllable image generation.
\item We proposed a semantic clothing encoder that injects multiple garment features into the denoising process through decoupled additive attention and applies attention reordering to balance the features of different garments.
\item 
Our model's performance, benchmarked against competitors, shows state-of-the-art results, highlighting the superiority of our approach.
\end{itemize}

\section*{2 RELATED WORK}
\subsection*{\raggedright 2.1 Controllable Image Generation}

In recent years, with the successful application of Latent Diffusion Models (LDMs) in text-to-image generation tasks, controllable image generation has gained significant attention. Many works have attempted to introduce spatial information into LDMs to enhance the controllability of the models. ControlNet is one of the most effective methods for controlling diffusion models, integrating a trainable copy of the UNet encoder block with zero initialization into the original UNet. T2I-Adapter proposes a compact network design that achieves the same functionality as ControlNet but with a lighter model. Uni-ControlNet introduces a unified framework that allows for flexible combinations of different conditional controls within a single model, further reducing training costs. However, the spatial structure information introduced by these methods only serves as auxiliary in the generation of model images and does not meet the requirements for producing customized images.
\subsection*{\raggedright 2.2 Subject-driven Human Image Synthesis}
Subject-driven human image synthesis aims to generate target subject's images using text prompts from given reference images, with the key challenge being to maintain high-fidelity visual features while generating new images in different contexts. This can be categorized into two main categories: test-time finetuning methods and finetuning-free methods. Test-time finetuning methods can produce satisfactory results but require more time to adapt to each aspect when customizing images with multiple garments. Generally, finetuning-free methods encode reference images into embeddings or image prompts without requiring additional finetuning. Thus, finetuning-free methods offer great flexibility and have more potential for practical applications. ELITE proposes global and local mapping schemes but suffers from limited fidelity. Instantbooth employs an adapter structure trained on domain-specific data to achieve subject-driven generation without finetuning. IP-Adapter designs a decoupled cross-attention mechanism that encodes images into prompts separate from the cross-attention layers of text features, achieving comparable performance to dedicated finetuned models in a simple manner. Versatile Diffusion extends existing single-stream diffusion pipelines into a multitask multimodal network to achieve cross-modal versatility. Magic Clothing uses a Reference UNet to extract features and implements feature fusion in the self-attention layer, but this method is limited to single garments. In this work, we focus on multi-garment-driven portrait synthesis, which requires the model to have the capability to retain details across multiple garments, create according to diverse text prompts, and be applicable to out-of-domain data without finetuning.

\section*{ 3 METHOD}

\subsection*{\raggedright 3.1 Preliminary}
Latent Diffusion Models (LDMs)have been successfully applied to text-to-image generation tasks. To add spatial conditioning control to pre-trained LDMs, ControlNetintegrates trainable encoder blocks into the original UNet architecture. Meanwhile, T2I-Adapterproposes a compact network design that offers the same functionality as ControlNet but with lower computational complexity. To further reduce training costs, Uni-ControlNetintroduces a unified framework that can handle different conditional controls flexibly and in combination within a single model.

On the other hand, LDMs have also played a significant role in image editing. InstructPix2Pix re-trains the UNet of LDMs by adding extra input channels to the first convolutional layer and fine-tuning it on a large dataset of image editing examples, enabling the model to follow editing instructions. MasaCtrl transforms self-attention in diffusion models into mutual self-attention, achieving consistent image generation and complex non-rigid image editing without additional training costs. InfEdit  achieves coherent and faithful semantic changes, both rigid and non-rigid, through denoising diffusion consistency models and attention control mechanisms.

In this paper, we leverage the powerful capabilities of pre-trained LDMs in text-to-image generation and introduce an additional clothing extractor to achieve clothing-driven image synthesis.
\subsection*{\raggedright 3.2 Garment Encoder}
We propose a semantic-aware garment encoder that identifies the category of the extracted features based on pre-specified garment categories as prompts, enabling effective matching of input garments to body parts.

We also utilize classifier-free guidance for multiple garments, which helps the diffusion model achieve a balance between sample quality and diversity through joint training of conditional and unconditional models.

Our model is designed as a plug-and-play module that can be combined with IP-Adapter and ControlNet, and it supports LoRA fine-tuning for base adapter compatibility.

Our training strategy involves pre-training on a single-garment subset of the DressCode dataset, followed by training on a multi-garment dataset.

\subsection*{\raggedright 3.3 Attention fusion Block }
\subsection*{\raggedright 3.3.1 Naive Settings}
We simply concatenate the garment features with the noise features before performing self-attention operations. This process causes the information from different garments to influence each other, failing to fully exploit the garment information.
\subsection*{\raggedright 3.3.2 Attention Concatenate}
We only concatenate the garment features with the noise features for the keys (K) and values (V), then compute the operation using the noise query (Q) with the concatenated K and V. During this process, the multiple garments still influence each other to some extent, such as after the Q and K operations.

\subsection*{\raggedright 3.3.3 Attention Addition}
We do not concatenate the garment features with the noise features; instead, we separately compute the operations between the noise features and the keys (K) and values (V) of each garment, and finally sum them up. This process completely decouples the attention computations for different garments, making it the best-performing method among the three.
\begin{figure*}[htb]
\centering
\label{figure}
\end{figure*}



\section*{4 EXPERIMENTS}
\subsection*{\raggedright Dataset}
All experiments were conducted on DressCode, a multi-garment dataset that was created by splitting additional garments from a single-garment dataset, forming triplet data pairs.
\subsection*{\raggedright Implementation Details}
In our experiments, we initialized the weights of our garment extractor by inheriting the pre-trained weights of the UNet component from Stable Diffusion v1.5. We fine-tuned only these weights while keeping the weights of the other modules unchanged. Our model underwent pre-training on single-garment data and subsequent fine-tuning on multi-garment data from the DressCode dataset, which has an image resolution of 768 × 576. Corresponding captions for the images were obtained from the BLIP~\cite{li2024blip} model. We utilized an AdamW optimizer with a fixed learning rate of 5e-5. The model was trained for 180 epochs with a batch size of 6 on a single NVIDIA V100 GPU. During inference, we employed the UniPC sampler for 20 sampling steps to generate images.

\subsection*{\raggedright Result comparison}
It can be observed that both the naive settings and the Attention Concatenate methods fail to adequately preserve the detailed information of multiple clothing items. In contrast, the Attention Addition method yields the best results. Furthermore, our approach, when combined with text prompts, enables the generation of customized avatars.\\

\section*{\raggedright CONCLUSION}
In conclusion, this paper presents Multi-Garment Customized Model Generation, a novel unified framework based on Latent Diffusion Models (LDMs) designed to synthesize images with free combinations of multiple pieces of clothing. By addressing the challenges of maintaining natural appearances and complex textures while preventing information interference, our method introduces a trainable garment encoder and decoupled multi-garment feature fusion. These innovations enable the parallel extraction of detailed garment features and the conditional generation of multiple garments, significantly reducing conflicts between garment information. The garment encoder is a flexible and extendable module that can integrate with other components like IP-Adapter and ControlNet, enhancing the diversity and controllability of generated images. Extensive experiments validate the superiority of our approach over existing methods, paving the way for advanced applications in the field of multi-garment image synthesis.


\bibliography{aaai24}

\end{document}